\documentclass[sigconf, nonacm]{acmart}

\newcommand\vldbdoi{XX.XX/XXX.XX}
\newcommand\vldbpages{XXX-XXX}
\newcommand\vldbvolume{20}
\newcommand\vldbissue{1}
\newcommand\vldbyear{2027}
\newcommand\vldbauthors{\authors}
\newcommand\vldbtitle{\shorttitle} 
\newcommand\vldbavailabilityurl{https://github.com/uiuc-kang-lab/ELT-Bench/pull/18}
\newcommand\vldbpagestyle{plain}

\usepackage{multirow}
\usepackage{makecell}
\usepackage{enumitem}

\begin{document}
\title{ELT-Bench-Verified:\\
Benchmark Quality Issues Underestimate AI Agent Capabilities}

\author{Christopher Zanoli}
\affiliation{%
  \institution{IBM Research \and ETH Zurich}
  \city{Zurich}
  \state{Switzerland}
}
\email{Christopher.Zanoli@ibm.com}

\author{Andrea Giovannini}
\affiliation{%
  \institution{IBM Research}
  \city{Zurich}
  \state{Switzerland}
}
\email{AGV@zurich.ibm.com}

\author{Tengjun Jin}
\affiliation{%
  \institution{University of Illinois (UIUC)}
  \city{Urbana}
  \state{USA}
}
\email{tengjun2@illinois.edu}

\author{Ana Klimovic}
\affiliation{%
  \institution{ETH Zurich}
  \city{Zurich}
  \state{Switzerland}
}
\email{aklimovic@ethz.ch}

\author{Yotam Perlitz}
\authornote{Corresponding author.}
\affiliation{%
  \institution{IBM Research}
  \city{Zurich}
  \state{Switzerland}
}
\email{Y.Perlitz@ibm.com}






\begin{abstract}
Constructing Extract-Load-Transform (ELT) pipelines---the backbone of modern data integration---remains one of the most labor-intensive tasks in data engineering, making it a high-impact target for AI-driven automation. On ELT-Bench, the first and only benchmark for evaluating AI agents on end-to-end ELT pipeline construction, agents achieved low success rates that suggested they are far from practical utility.
We revisit these results and identify two factors that led to a substantial underestimation of agent capabilities.
First, re-evaluating ELT-Bench with the same agent framework but upgrading only the underlying large language model reveals that the extraction and loading stage is largely solved, while transformation performance improves dramatically.
Second, we develop an \textit{Auditor-Corrector} methodology that combines scalable LLM-driven root-cause analysis with rigorous human validation (inter-annotator agreement Fleiss' $\kappa = 0.85$) to systematically audit benchmark quality. Applying this methodology to ELT-Bench uncovers that the majority of failed transformation tasks contain benchmark-attributable errors---including overly rigid evaluation scripts, ambiguous specifications, and incorrect ground truth---that penalize correct agent outputs.
Based on these findings, we construct ELT-Bench-Verified, a revised benchmark with refined evaluation logic and substantial ground-truth revisioning.
Re-evaluating on ELT-Bench-Verified yields a significant further improvement attributable entirely to benchmark correction.
Our results demonstrate that both rapid model improvement and benchmark quality issues contributed to a substantial underestimation of agent capabilities in the original evaluation. More broadly, our findings echo recent observations of pervasive annotation errors in text-to-SQL benchmarks, suggesting that benchmark quality issues are a systemic problem across data engineering evaluation. Systematic quality auditing should become standard practice, especially for benchmarks that evaluate complex, multi-step agentic tasks. We release ELT-Bench-Verified as a community resource to provide a more reliable foundation for measuring and driving further progress in AI-driven data engineering automation.
\end{abstract}

\maketitle

\pagestyle{\vldbpagestyle}
\begingroup\small\noindent\raggedright\textbf{PVLDB Reference Format:}\\
\vldbauthors. \vldbtitle. PVLDB, \vldbvolume(\vldbissue): \vldbpages, \vldbyear.\\
\href{https://doi.org/\vldbdoi}{doi:\vldbdoi}
\endgroup
\begingroup
\renewcommand\thefootnote{}\footnote{\noindent
This work is licensed under the Creative Commons BY-NC-ND 4.0 International License. Visit \url{https://creativecommons.org/licenses/by-nc-nd/4.0/} to view a copy of this license. For any use beyond those covered by this license, obtain permission by emailing \href{mailto:info@vldb.org}{info@vldb.org}. Copyright is held by the owner/author(s). Publication rights licensed to the VLDB Endowment. \\
\raggedright Proceedings of the VLDB Endowment, Vol. \vldbvolume, No. \vldbissue\ %
ISSN 2150-8097. \\
\href{https://doi.org/\vldbdoi}{doi:\vldbdoi} \\
}\addtocounter{footnote}{-1}\endgroup

\ifdefempty{\vldbavailabilityurl}{}{
\vspace{.3cm}
\begingroup\small\noindent\raggedright\textbf{PVLDB Artifact Availability:}\\
The source code, data, and/or other artifacts have been made available at \url{\vldbavailabilityurl}.
\endgroup
}

\begin{figure}[t!]
  \vspace{0.5cm}
  \centering
  \includegraphics[width=\linewidth]{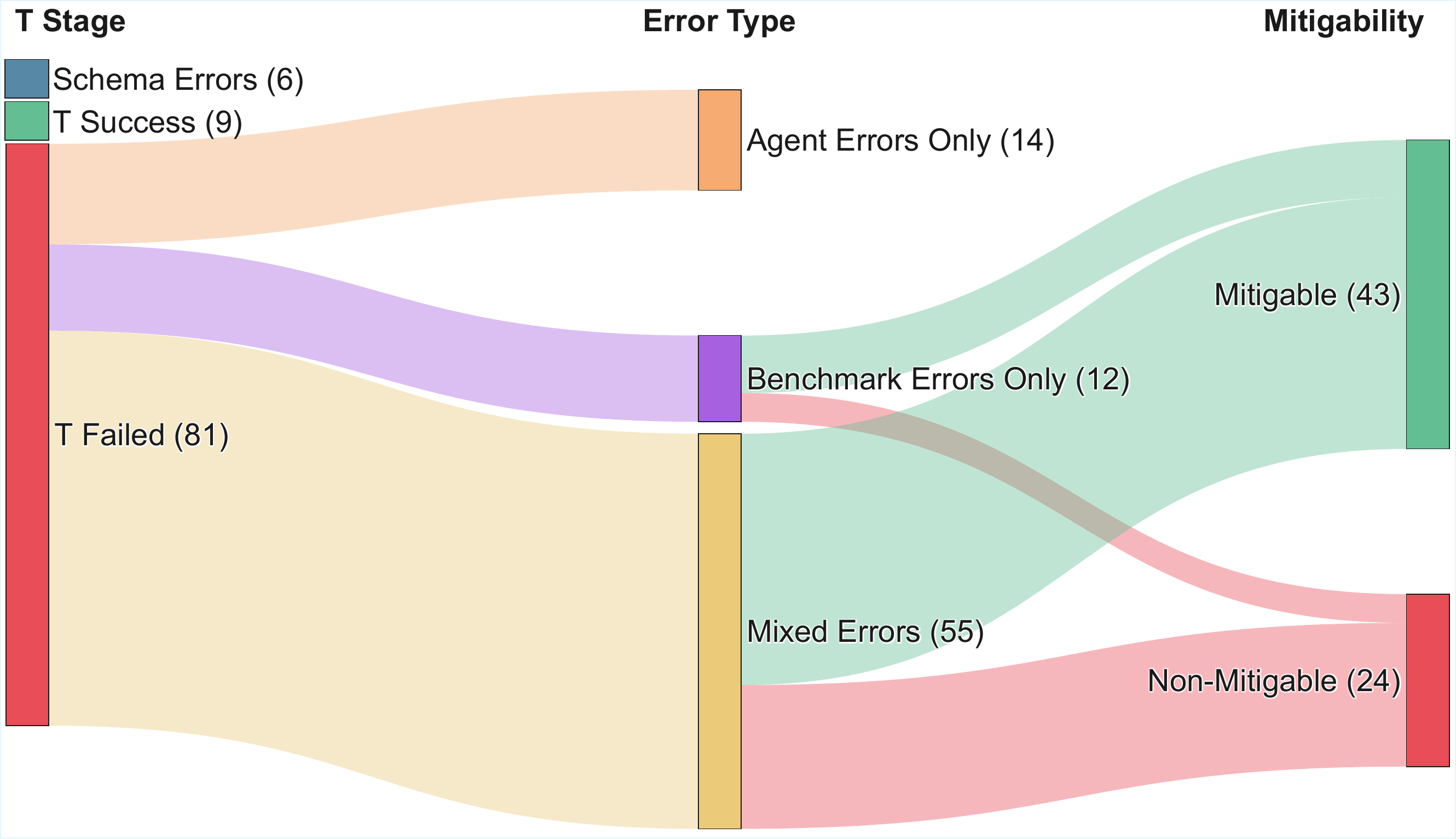}
  \caption{Distribution of error attribution across 81 failed transformation tasks in ELT-Bench. Tasks are classified by error source---agent-attributable, benchmark-attributable, or mixed---and further stratified by mitigability, distinguishing errors addressable through evaluation refinements from those requiring ground truth column removal.}
  \label{fig:cover}
\end{figure}

\section{Introduction}

\begin{figure}[tb]
  \centering
  \includegraphics[width=\linewidth]{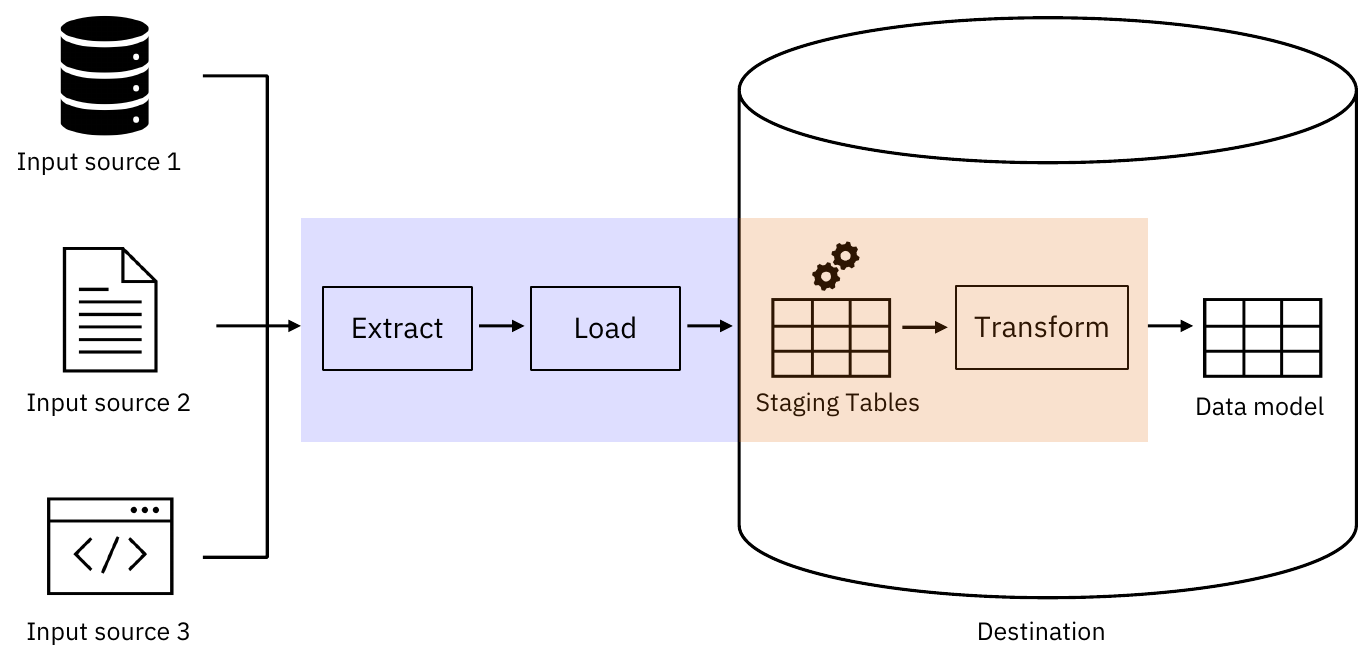}
  \caption{Overview of a typical ELT pipeline. Data is first \textbf{extracted} from heterogeneous sources---such as databases, APIs, flat files---using connectors like Airbyte~\cite{airbyte}. It is then \textbf{loaded} in raw form into a cloud data warehouse such as Snowflake~\cite{snowflake}. Finally, \textbf{transformation} logic, typically written as SQL models in dbt~\cite{dbt}, reshapes the raw data into analytics-ready tables within the warehouse.}
  \label{fig:elt_overview}
\end{figure}

Modern organizations rely heavily on Extract-Load-Transform (ELT) pipelines---workflows that extract data from heterogeneous sources, load it into a cloud warehouse, and transform it in place (Figure~\ref{fig:elt_overview})---to integrate and analyze data at scale~\cite{frometltoelt}. 

However, constructing these pipelines remains a highly manual, labor-intensive process requiring expertise across diverse data sources~\cite{eltpipelineseffort1, eltpipelineseffort2}, cloud warehouses like Snowflake~\cite{snowflake}, and transformation frameworks like dbt~\cite{dbt}. 
The prospect of automating this workflow with AI agents is a major goal for data engineering. 
To measure progress toward this goal, ELT-Bench~\cite{eltbench} was recently introduced as the first benchmark for end-to-end ELT pipeline construction. 
Yet, the initial baseline results were stark: SWE-Agent with Claude Sonnet 3.5~\cite{claude35sonnet} achieved only a $37\%$ success rate on data extraction and loading and a mere $1\%$ on data transformation. 
These figures suggested that current AI agents are far from practical utility in real-world data engineering tasks.

We revisit these findings and show that this assessment was premature, driven by two compounding factors.
First, the original evaluation captured a snapshot of older capabilities. 
By upgrading only the underlying model to Claude Sonnet 4.5~\cite{claude45sonnet} while keeping the same agent framework (SWE-Agent~\cite{swe-agent}), we observe a substantial leap in performance: extraction and loading rises from 37\% to 96\%---indicating that this phase is largely solved for the source types and tool configurations covered by ELT-Bench---while the transformation success rate jumps from 1\% to 22.66\%.

Despite this substantial improvement, a 22.66\% transformation success rate appears surprisingly low given that modern approaches to related tasks such as text-to-SQL generation now exceed 81\% execution accuracy~\cite{shkapenyuk2025text2sql, wang2025agentarscalesql}. This gap led us to question whether we had reached the ceiling of the agent's capabilities, or the ceiling of the benchmark's validity.
Motivated by recent discoveries of pervasive annotation errors in text-to-SQL benchmarks~\cite{jin2025pervasive}, we conduct a systematic data quality audit of ELT-Bench. Of the 100 benchmark tasks, 96 pass extraction and loading with the upgraded model; of those, 87 fail the transformation stage, and 81 produce transformation outputs yet fail column-level evaluation---these 81 tasks form the scope of our audit (Figure~\ref{fig:simple}). Through a rigorous process validated by three independent annotators (Fleiss' $\kappa = 0.85$), our investigation uncovers pervasive evaluation issues: 82.7\% of these 81 tasks contain at least one benchmark-attributable error. At a finer granularity, each task comprises multiple columns that are evaluated independently; at this column level, 33.0\% of all mismatches are benchmark-attributable rather than genuine agent failures. These errors range from ambiguous task specifications and overly rigid evaluation scripts to mathematically incorrect ground-truth values (Table~\ref{tab:error_distribution}, Figure~\ref{fig:cover}).

To establish a reliable measure of AI capabilities in data engineering, we introduce a two-stage \textbf{Auditor-Corrector} framework (Figure~\ref{fig:method})---a general, human-validated methodology for auditing complex data engineering benchmarks at scale. Our \textit{Auditor} pipeline combines LLM-driven root-cause analysis with comprehensive manual verification: it systematically reverse-engineers the correct SQL for each unmatched column and classifies every mismatch by attribution, with all findings reviewed by domain experts and validated through an inter-annotator agreement study (Fleiss' $\kappa = 0.85$). Our \textit{Corrector} pipeline then translates these findings into targeted benchmark refinements to construct ELT-Bench-Verified. By re-evaluating the exact same agent and model on this verified benchmark, the transformation success rate rises from 22.66\% to 32.51\%---a 43.5\% relative improvement attributable entirely to benchmark correction. In summary, our core contributions are:

\begin{enumerate}[leftmargin=*]
    \item \textbf{Updated EL Stage Baseline:} We show that extraction and loading is largely solved for ELT-Bench's covered configurations, with SRDEL rising from 37\% to 96\% by upgrading only the underlying model.
    \item \textbf{An Auditor-Corrector Methodology:} We develop a scalable methodology for auditing execution-based benchmarks, combining LLM-driven root-cause analysis with structured human verification. Applied to ELT-Bench, it classifies 660 column-level mismatches into a novel error taxonomy, revealing that 82.7\% of failed tasks contain benchmark-attributable errors.
    \item \textbf{ELT-Bench-Verified:} We construct a corrected benchmark by refining evaluation scripts and removing unreliable ground-truth columns, raising SRDT from 22.66\% to 32.51\% with the same agent and model, and release it as a community resource for reliable evaluation of AI agents on ELT tasks.
\end{enumerate}

\section{Revisiting ELT-Bench}
\label{sec:reeval}

\begin{figure*}[t]
    \centering
    \includegraphics[width=\textwidth]{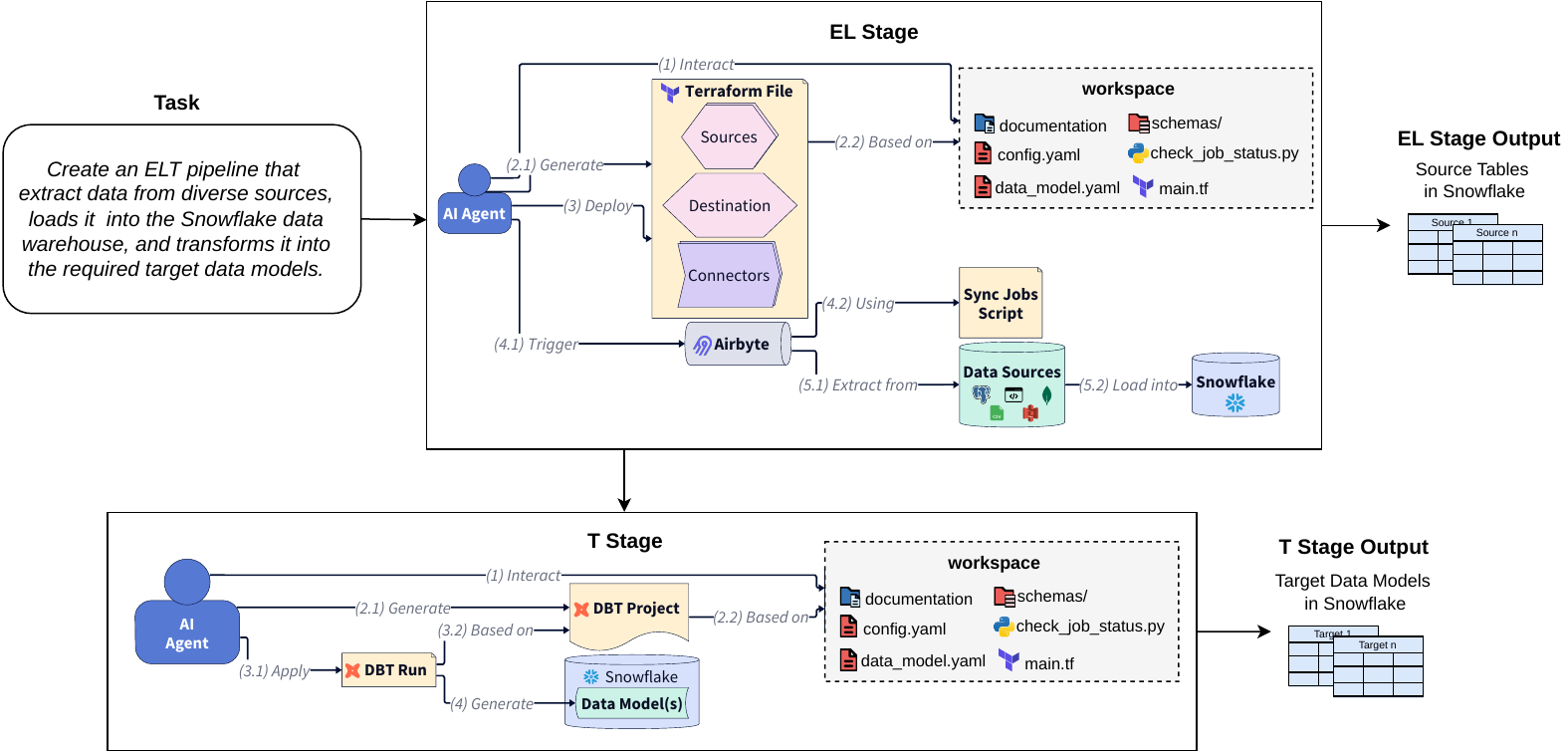}
    \caption{Agent workflow in ELT-Bench, illustrating the two-stage pipeline the agent must construct. \textbf{EL Stage:} The agent reads connection details from \texttt{config.yaml}, writes Terraform configurations to define Airbyte~\cite{airbyte} sources, the Snowflake~\cite{snowflake} destination, and their connectors, deploys the configuration, and triggers synchronization jobs that extract data from heterogeneous sources (databases, APIs, flat files) and load it into the warehouse as staging tables. The agent then uses a provided Python script in the workspace to poll job status and proceeds to the T stage once all jobs succeed (i.e., all source tables for the task have been loaded into Snowflake). \textbf{T Stage:} The agent initializes a dbt~\cite{dbt} project, authors SQL transformation models that reshape the staged data into the target data models defined in \texttt{data\_model\_schema.yaml}, and executes \texttt{dbt run}, which generates the target data model(s) in the Snowflake data warehouse.}
    \label{fig:agent_flows}
\end{figure*}

ELT-Bench~\cite{eltbench} is the first benchmark for evaluating AI agents on end-to-end ELT pipeline construction. It comprises 100 tasks spanning diverse databases, each requiring the agent to build a complete pipeline from a provided starter codebase containing connection configurations, target data model specifications, source table schemas, and data tool documentation. Figure~\ref{fig:agent_flows} illustrates the two-stage agent workflow. Performance is measured by two metrics: SRDEL (Success Rate for Data Extraction \& Loading), the fraction of tasks where all source data is correctly loaded, and SRDT (Success Rate for Data Transformation), the fraction of total target data models whose columns match the ground truth.

Before auditing benchmark quality, we first establish an updated performance baseline by re-evaluating ELT-Bench with a more recent model than those used in the original study.

\subsection{Experimental Setup}

We evaluate SWE-Agent~\cite{swe-agent} with Claude Sonnet 4.5 across all 100 ELT-Bench tasks, using the same evaluation protocol and metrics (SRDEL and SRDT) defined by the original authors. We select SWE-Agent because it was one of two agent frameworks evaluated in the original ELT-Bench study, enabling direct comparison. We focus on a single model upgrade rather than evaluating multiple model--agent combinations due to the substantial computational cost of each full evaluation run: a single SWE-Agent run across all 100 tasks requires 2 days, 7 hours and 16 minutes of wall-clock time and costs \$343 in API fees, while a Baseline (ReAct) run requires 18 hours and 17 minutes at \$293. The cumulative cost of the two runs reported in this paper alone exceeds 3 days and \$635.

\subsection{Results}

\begin{figure}[tb]
\centering
\includegraphics[width=\linewidth]{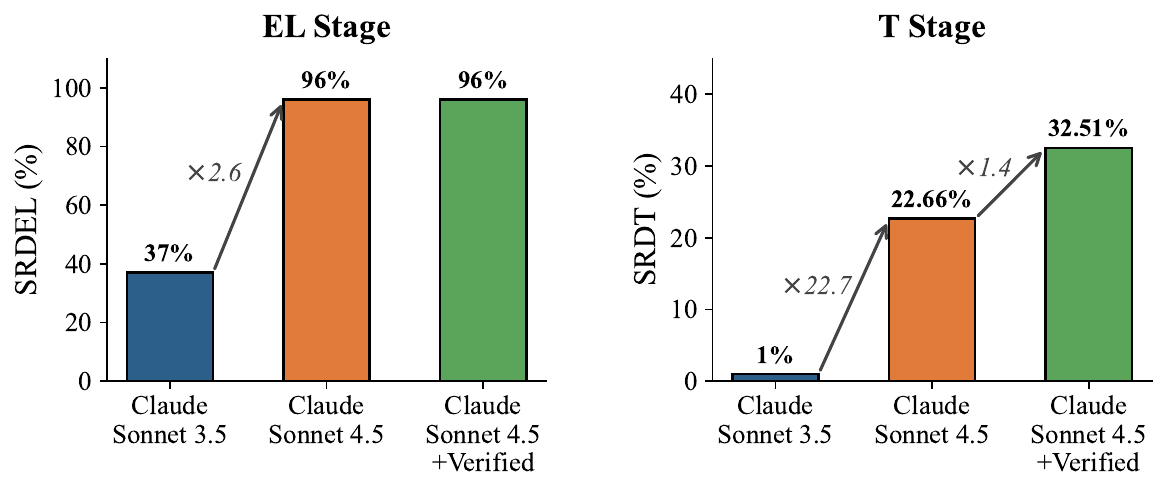}
\caption{Performance progression across three configurations, all using SWE-Agent: the original model (Claude Sonnet 3.5), the upgraded model (Claude Sonnet 4.5), and the upgraded model evaluated on ELT-Bench-Verified. The EL stage is largely solved after the model upgrade ($\times$2.6), while the T stage benefits from both model improvement ($\times$22.7) and benchmark correction ($\times$1.4).}
\label{fig:re_evaluation}
\end{figure}

Figure~\ref{fig:re_evaluation} presents our results alongside the original ELT-Bench findings, including the effect of benchmark correction (ELT-Bench-Verified, detailed in Section~\ref{sec:verified}). In terms of data models passed, the original model achieves 2/203, the upgraded model 46/203, and the upgraded model on ELT-Bench-Verified 66/203.

The data extraction and loading stage shows a 59 percentage point improvement (37\% to 96\%), with 96 of 100 tasks successfully loading data into Snowflake. Since the agent framework is held constant, this improvement is attributable to the model upgrade. This result suggests that, for the source types and tool configurations covered by ELT-Bench, the extraction and loading stage is largely solved.

The transformation stage improves from 1\% to 22.66\% SRDT. Of the 96 successfully loaded tasks, 9 pass the transformation stage (accounting for 46 of 203 data models), 87 fail, and 6 of those 87 fail due to schema errors---non-existent tables caused by premature agent termination---leaving 81 tasks that produce transformation outputs but fail column-level evaluation. Including the 4 tasks that failed extraction entirely, 77.3\% of all data models (157/203) do not pass transformation evaluation.

This raises a natural question: to what extent do these remaining failures reflect genuine agent limitations versus benchmark quality issues? We investigate this in the following sections.

\begin{figure}[tb]
  \centering
  \includegraphics[width=0.5\textwidth]{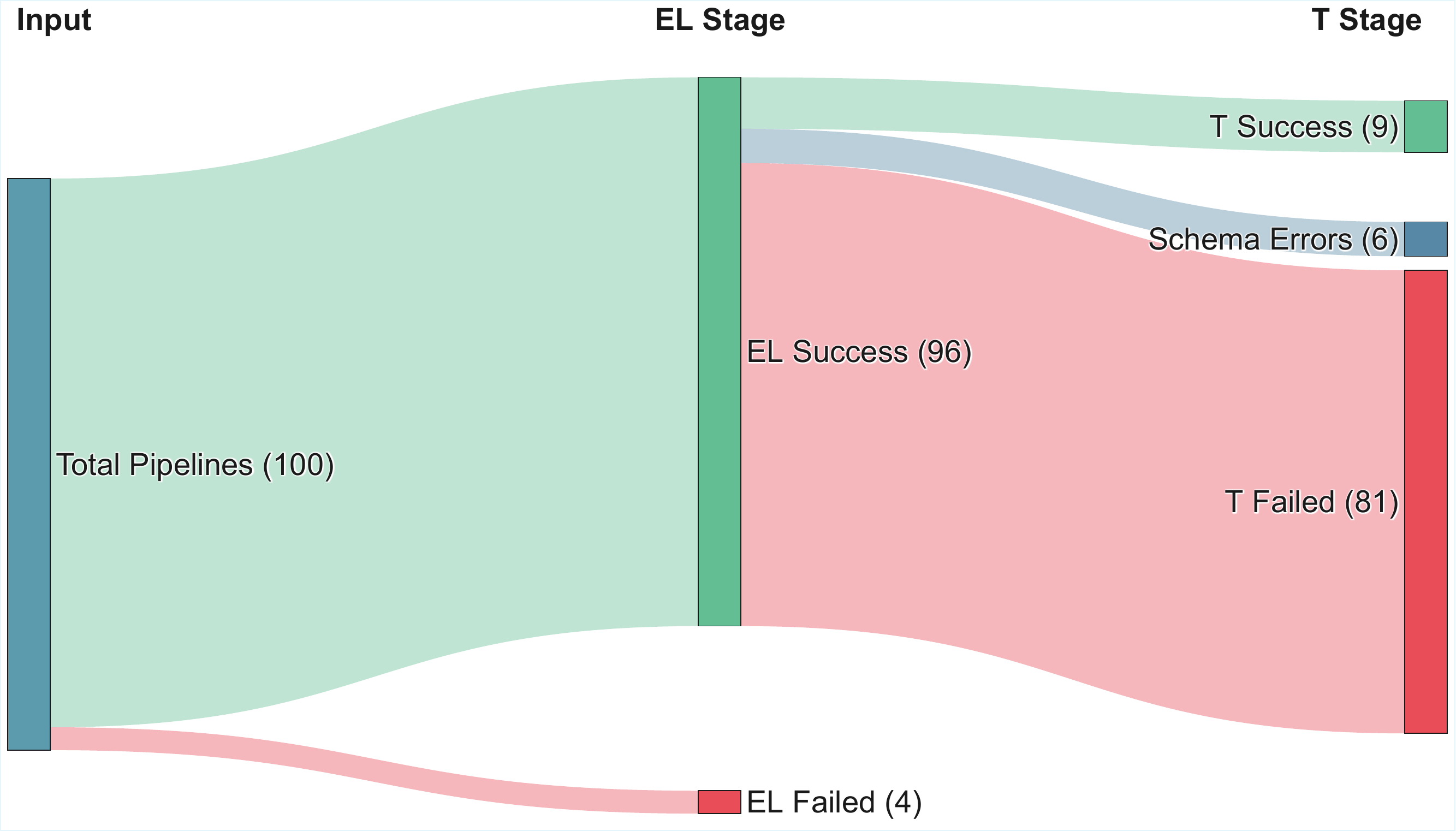}
  \caption{Pipeline outcomes across extraction/loading and transformation stages.}
  \label{fig:simple}
\end{figure}

\section{Methodology}
\label{sec:methodology}

\begin{figure*}[t]
    \centering
    \includegraphics[width=\textwidth]{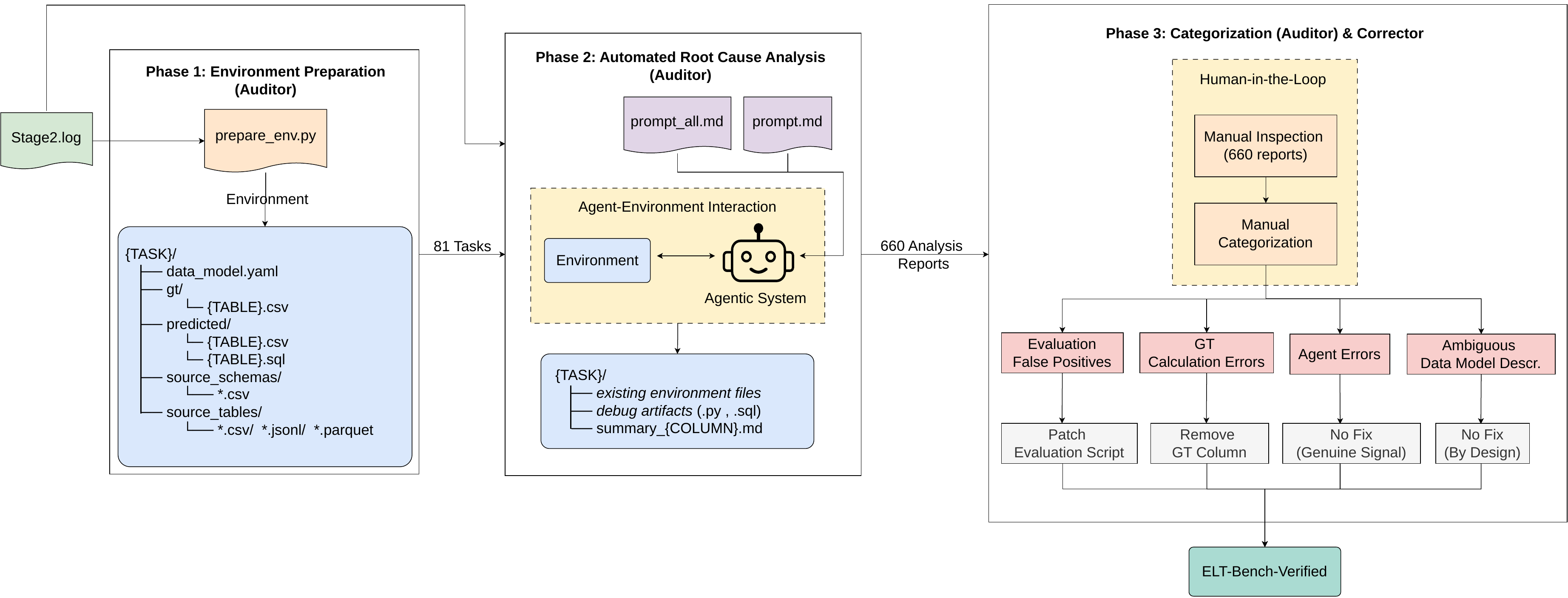}
    \caption{Overview of the Auditor-Corrector framework. The \textit{Auditor} proceeds in three phases: Phase~1 constructs a structured analysis environment for each failed task from the evaluation log; Phase~2 deploys an LLM agent per task that autonomously reverse-engineers the correct SQL for each unmatched column, verifying each derivation against the ground truth; Phase~3 involves manual categorization of all 660 reports into an error taxonomy. The \textit{Corrector} then applies category-specific corrections---evaluation script refinements and removal of columns with unreliable ground truth---to produce ELT-Bench-Verified.}
    \label{fig:method}
\end{figure*}

To audit and refine ELT-Bench, we introduce a two-stage \textit{Auditor-Corrector} framework. The Auditor reverse-engineers the correct SQL for each unmatched column and classifies every mismatch by attribution; the Corrector translates these findings into targeted benchmark refinements. Figure~\ref{fig:method} provides an overview.

\paragraph{Formal setup.} ELT-Bench comprises $N$ database tasks $d_1, \ldots, d_N$. Each task $d_i$ contains one or more target data models, and each data model defines a set of target columns. We write $\mathcal{C}_i$ for all columns in task $d_i$. The evaluation script compares each predicted column against its ground truth counterpart and marks it as \emph{matched} or \emph{unmatched}. We define $\mathcal{U} \subseteq \bigcup_{i} \mathcal{C}_{i}$ as the set of all unmatched columns across all tasks. Our framework operates exclusively over $\mathcal{U}$. In our instantiation, $N = 81$ tasks (those that produced transformation outputs but failed evaluation), yielding $|\mathcal{U}| = 660$ unmatched columns distributed across 136 data models.

\subsection{The Auditor Pipeline}
The Auditor combines automated analysis with systematic manual verification to reverse-engineer the correct SQL for each unmatched column and classify every mismatch. It proceeds in three phases.

\subsubsection{Phase 1: Analysis Environment Preparation}
\label{sec:phase1}

For each task $d_i$ containing at least one unmatched column, we construct a structured analysis environment populated with the artifacts necessary for root cause analysis: the task specification (\texttt{data\_model.yaml}), source tables and schemas, ground truth tables\\(\texttt{gt/<data\_model>.csv}), and agent-generated SQL and predictions\\(\texttt{predicted/<data\_model>.sql}, \space \texttt{predicted/<data\_model>.csv}). The environment is populated selectively: tasks where all columns match are excluded entirely, and within each task, only data models containing at least one unmatched column are included. This filtering reduces the analysis scope from the full benchmark ($N = 100$ tasks, 203 data models) to the relevant subset ($N = 81$ tasks, 136 data models, 660 columns).

\subsubsection{Phase 2: Scaled Analysis via LLM Agent}
\label{sec:phase2}

We adopt an agentic approach in which an LLM-powered agent (Claude Opus 4.5~\cite{claude45opus}) is instantiated within each task's analysis environment and tasked with diagnosing every column-level mismatch. Analysis is parallelized at the task level: one agent instance is spawned per task $d_i$, with full read access to all task artifacts. The agent is guided by a two-level prompt architecture: an \emph{orchestration prompt}\\ (\texttt{prompt\_all.md}) parses the evaluation log, builds a work queue of (database, table, columns) tuples, and dispatches each item to an \emph{analysis prompt} (\texttt{prompt.md}) that encodes the per-column analytical protocol. Both prompts are released as supplementary material.

For each unmatched column $c \in \mathcal{U} \cap \mathcal{C}_i$, the agent autonomously navigates the task environment---inspecting the column specification, tracing the relevant SQL logic in the agent's predicted query, comparing predicted and ground truth values, and reasoning over the observed discrepancies. The agent iteratively formulates hypotheses about the mismatch root cause and tests them by deriving alternative SQL interpretations and executing each against the source data to compare with the ground truth. For each column, the agent produces a structured analysis report containing: (i)~a root cause diagnosis identifying the specific flaw; (ii)~a corrected SQL derivation, \emph{verified} by the agent to achieve a 100\% row-level match with the ground truth when executed against the source data; and (iii)~supporting evidence including value-level comparisons and match statistics demonstrating the exact match. For the 30 columns where no SQL interpretation achieves a perfect match---indicating that the ground truth values cannot be derived from any reasonable reading of the task specification---the agent explicitly flags the column and reports the closest approximation. In total, this phase produces 660 analysis reports covering all unmatched columns.

\subsubsection{Phase 3: Categorization and Correction}
\label{sec:phase3}

A data engineer manually inspects all 660 analysis reports. For each report, the reviewer verifies that (i)~the identified root cause is consistent with the presented evidence, (ii)~the corrected SQL logic (if proposed) is semantically sound, and (iii)~the match statistics are accurate.

Following inspection, the reviewer assigns each column to an error category. The resulting categorization assigns each column $c \in \mathcal{U}$ to exactly one error category along two dimensions: \emph{attribution} (agent-attributable vs.\ benchmark-attributable) and \emph{mitigability} (whether the error can be corrected without modifying ground truth data). The primary categorization is performed by a single data engineer. To ensure reproducibility, three independent annotators classified a uniformly sampled subset of 50 columns using the full 14-category taxonomy, given only the analysis reports and a detailed annotation protocol. The resulting agreement is strong at both granularity levels: for the high-level attribution (agent vs.\ benchmark), Fleiss' $\kappa$ reaches 0.851 (almost perfect, 94.7\% agreement), and for the exact 14-category classification, Fleiss' $\kappa$ reaches 0.755 (substantial, 79.3\% agreement), confirming that the categorization is reproducible across annotators.

\subsubsection{Validation of the Auditor}

The structured prompt instructs the agent to produce, for each unmatched column, a dedicated analysis report and a Python debug script. Because the agent operates autonomously, its intermediate artifacts exhibit natural variation: debug scripts were sometimes written in SQL rather than Python (9 of 81~tasks), or produced in multiple iterative versions as the agent refined its hypotheses (27~tasks). In 4~tasks where a single systemic root cause affected all columns in a table, the agent produced a combined analysis report rather than separate per-column reports. Despite these deviations, every unmatched column is covered: the 660 columns are addressed across 528 analysis reports, with multi-column reports accounting for the difference.

Crucially, for 630 of 660 columns the agent's corrected SQL is self-validated: the agent programmatically executes the corrected query against the source data and confirms a 100\% row-level match with the ground truth before finalizing the report. This execution-based verification ensures that each diagnosis is grounded in empirical evidence rather than reasoning alone. For the remaining 30 columns, the agent explicitly reports that no SQL interpretation achieves a perfect match, flagging them as potential ground truth errors---a finding later confirmed by the independent validation study described in Section~\ref{sec:corrector_validation}. A data engineer subsequently reviewed all 660 reports, verifying the root cause identification, corrected SQL derivation, and supporting evidence in each case; no report required correction of its core diagnostic content.

\subsection{The Corrector Pipeline}
\label{sec:corrector}
The Corrector translates the Auditor's findings into benchmark refinements without introducing new biases.

\subsubsection{Correction Logic}

The categorization from Phase~3 classifies each column along two dimensions: \emph{attribution} (benchmark-attributable vs.\ agent-attributable) and \emph{mitigability}. Benchmark-attributable errors fall into three categories: \emph{Evaluation False Positives} (correct agent outputs penalized by rigid evaluation scripts), \emph{Ground Truth Calculation Errors} (ground truth values that cannot be derived from any reasonable interpretation), and \emph{Ambiguous Data Model Descriptions} (underspecified column definitions). The full taxonomy, including agent-attributable error categories, is presented in Section~\ref{sec:taxonomy}. Based on this categorization, we apply three types of corrections:
\begin{itemize}
    \item \textbf{Evaluation Script Refinements:} For columns classified as Evaluation False Positives, we patch the evaluation script to handle representational equivalences (e.g., boolean normalization, floating-point tolerance, percentage format normalization, NULL equivalence, order-insensitive comparison). Each patch is derived directly from the mismatch pattern documented in the analysis report.
    \item \textbf{Ground Truth Column Removal:} For columns classified as Ground Truth Calculation Errors---where no SQL interpretation achieved a 100\% match during Phase~2---we remove the affected columns from the benchmark. An independent validation study (Section~\ref{sec:corrector_validation}) confirms that no reliable correction can be established for these columns, as human experts themselves disagree on the correct interpretation. Only the 30 affected columns are removed; all remaining columns in each data model are preserved.
    \item \textbf{Preservation of Ambiguity:} We deliberately preserve specification ambiguity for columns classified as Ambiguous Data Model Descriptions. Handling real-world uncertainty is a capability that agents should possess, and this aspect of the benchmark should be evaluated rather than simplified.
\end{itemize}

Agent-attributable errors receive no correction, as these represent genuine evaluation signal.

\subsubsection{Validation of the Corrector}
\label{sec:corrector_validation}
We validate the Corrector along two axes, corresponding to the two types of corrections applied.

\paragraph{Evaluation Script Refinements.}
To verify that the patched evaluation script aligns with human judgment better than the original, we conduct a stratified annotation study. We run both the original and patched evaluation scripts on the same agent outputs (SWE-Agent with Claude Sonnet 4.5) using the original ground truth, then classify each column into one of three strata based on the two scripts' labels: (\textbf{A})~both scripts report a match, (\textbf{B})~the original script reports a mismatch but the patched script reports a match (i.e., corrected false positives), and (\textbf{C})~both scripts report a mismatch. We exclude columns associated with tasks that failed during the extraction stage, as well as the 30 columns identified with ground truth errors (i.e., those classified under the \textit{ground truth error} category in Phase 3 of our methodology). This filtering isolates the effect of the script refinements.

After excluding columns from tasks that failed extraction (4 tasks) or encountered schema errors (6 tasks), as well as the 30 ground truth error columns, the eligible population comprises 2{,}202 columns: 1{,}572 in Stratum~A, 156 in Stratum~B, and 474 in Stratum~C. We sample 50 columns using stratified sampling: 15 from Stratum~A, 20 from Stratum~B, and 15 from Stratum~C, deliberately oversampling Stratum~B (the corrected false positives) to ensure sufficient statistical power on the decisive stratum where the two scripts differ. For each sampled column, three engineers independently label each column pair as \emph{semantically equivalent} or \emph{not equivalent}, given the ground truth values, the agent's predicted values, and the column's natural language description, without knowledge of which stratum the column belongs to or what either script reported. We measure inter-annotator agreement using Fleiss'~$\kappa$, obtaining $\kappa = 0.664$ (substantial agreement), with 86\% of columns receiving a unanimous label. We resolve disagreements by majority vote and use the resulting consensus labels as ground truth for all subsequent analysis.

Stratum~B is the decisive stratum: these are the columns the patch reclassified from mismatch to match. We assess concordance between the consensus human labels and each script using McNemar's test with Yates' continuity correction throughout, which evaluates whether disagreements are symmetric or systematically biased in one direction. For Stratum~B, the original script disagrees with human judgment on all 20~columns, always in the same direction---labeling columns as mismatches that humans consider equivalent---a bias that is highly statistically significant (McNemar $\chi^2 = 18.05$, $p < 0.0001$). The patched script agrees with human judgment on all 20~Stratum~B columns, with no discordant pairs. For Stratum~A, both scripts achieve perfect concordance with human labels. For Stratum~C, neither script agrees with human consensus on any of the 15 columns. The discordant pairs are nearly balanced ($b=7$, $c=8$), yielding McNemar $\chi^2 = 0.00$ ($p = 1.0$) with continuity correction, indicating that neither script exhibits a systematic directional bias on these hard cases. The 0\% agreement on Stratum~C reflects the inherent difficulty of these cases: unlike the clear-cut representational equivalences corrected in Stratum~B (e.g., boolean normalization, format differences), Stratum~C columns involve subtler semantic judgments on which even human annotators do not fully agree. The fact that 7 of these 15 columns are labeled as equivalent by humans but as mismatches by both scripts suggests that our corrections are conservative by design---we address only the false positives for which strong consensus exists, and any residual cases would further strengthen the reported improvement. Over the stratified sample (which oversamples Stratum~B to maximize statistical power on the decisive stratum), the patched script agrees with human judgment on 70\% of sampled columns versus 30\% for the original; note that per-stratum results are more informative than these unweighted aggregates, and its remaining disagreements show no directional bias (McNemar $\chi^2 = 0.00$, $p = 1.0$), in contrast to the original script's strongly asymmetric error pattern (McNemar $\chi^2 = 10.31$, $p = 0.001$). These results confirm that the patch corrects genuine false positives validated by human judgment, without introducing new systematic errors.


\paragraph{Discarding Columns with Ground Truth Errors.}
During the error analysis phase, our auditor flagged 30 columns (spanning 24 tasks) as having incorrect ground truth and proposed revised SQL queries to correct them. Before accepting these revisions, we conduct an independent validation study to assess whether a reliable correction can be established. Three data engineers---with no access to the auditor's proposed SQL, the original ground truth, or any prior
analysis---independently author SQL queries for all 30 columns using only the task specification, source schemas, and source data. Each query is executed and results are compared pairwise among engineers and against the auditor's proposed values via exact match, after applying the same normalization rules used in ELT-Bench-Verified's patched evaluation script (boolean normalization, floating-point tolerance, format normalization, NULL equivalence, and order-insensitive comparison). This ensures the comparison is not artificially deflated by the representational equivalences identified earlier in our analysis.


Pairwise inter-annotator agreement ranges from 46.7\% to 70.0\%, with an average of 57.8\%. This level of disagreement among domain experts with full access to the source data indicates that no unambiguous correction exists for these columns.
Concordance with the auditor's proposed values is lower still, averaging 36.7\% across engineers. Majority-vote consensus is available for only 24 of the 30 columns; on the remaining 6 columns, all three engineers produced entirely different results, further underscoring the fundamental ambiguity of these cases. Among the 24 columns with a majority, the consensus agrees with the auditor on only 41.7\% of cases. The low inter-annotator agreement is the decisive criterion: it establishes that no single corrected ground truth can be asserted with confidence for these columns, rendering any revision arbitrary. Replacing the original erroneous values with the auditor's proposals would therefore not constitute a correction but merely a substitution of one unreliable value for another. We consequently discard all 30 columns from the benchmark while retaining all remaining columns belonging to the same data model, amounting to a removal of just 30 out of 2{,}494 columns (1.2\%) with no loss of data model coverage.

Additionally, we conduct an ablation study (Section~\ref{sec:ablation}) to disentangle the contributions of evaluation script refinements and ground truth column removal.

\section{Audit Findings}
\label{sec:taxonomy}

\begin{table}[t!]
\centering
\caption{Distribution of error categories across 660 unmatched columns.}
\label{tab:error_distribution}
\begin{tabular}{llrr}
\toprule
\textbf{Attribution} & \textbf{Error Category} & \textbf{Count} & \textbf{\%} \\
\midrule
\multirow{6}{*}{Agent} & Flawed SQL Logic & 221 & 33.5\% \\
& JOIN Type Errors & 157 & 23.8\% \\
& Wrong Data Source & 38 & 5.8\% \\
& Domain Knowledge Gaps & 16 & 2.4\% \\
& Key Generation Errors & 6 & 0.9\% \\
& NULL Handling Errors & 4 & 0.6\% \\
\cmidrule{2-4}
& \textit{Subtotal (Agent)} & \textit{442} & \textit{67.0\%} \\
\midrule
\multirow{3}{*}{Benchmark} & Evaluation False Positives & 156 & 23.6\% \\
& Ambiguous Data Model Descr. & 32 & 4.8\% \\
& GT Calculation Errors & 30 & 4.5\% \\
\cmidrule{2-4}
& \textit{Subtotal (Benchmark)} & \textit{218} & \textit{33.0\%} \\
\midrule
& \textbf{Total} & \textbf{660} & \textbf{100\%} \\
\bottomrule
\end{tabular}
\end{table}

Following our structured assessment, we present the quantitative and qualitative findings of our audit. We classify errors along two dimensions: attribution (benchmark vs.\ agent) and mitigability. We first present the full error taxonomy, then discuss the quantitative distribution.

\subsection{Error Taxonomy}

\subsubsection{Agent-Attributable Errors}
We define agent-attributable errors as discrepancies arising from flawed SQL logic, incorrect data source selection, or semantic misinterpretation. These represent genuine agent errors and should therefore be counted as failures. We categorize such errors into the following classes:

\paragraph{Flawed SQL Logic.}
Syntactically correct but logically flawed SQL, including: incorrect aggregation methods, erroneous formulas,\\ boolean logic inversions, incorrect window function specifications, and type mismatches.

\paragraph{JOIN Type Errors.}
Incorrect JOIN type selection causing missing or extraneous rows---typically \texttt{INNER JOIN} excluding valid rows or \texttt{FULL OUTER JOIN} creating spurious combinations.

\paragraph{Wrong Data Source.}
Queries using inappropriate source tables or columns, including missing required joins or selection of incorrect columns from correct tables.

\paragraph{Domain Knowledge Gaps.}
Lack of domain-specific understanding required for correct interpretation (e.g., ``winnings'' interpreted as race victories rather than points earned).

\paragraph{NULL Handling Errors.}
Incorrect NULL semantics, such as failing to use \texttt{NULLS FIRST} in ordering or misinterpreting implicit NULL meanings.

\paragraph{Key Generation Errors.}
Primary key generation using different methodology than ground truth, causing row matching failures.

\subsubsection{Benchmark-Attributable Errors}
We define benchmark-\\attributable errors as discrepancies arising from deficiencies in benchmark design rather than from agent reasoning. Specifically, these errors stem from ambiguous task specifications, limitations in the evaluation methodology, or incorrect ground truth values, leading to correct agent outputs being incorrectly classified as failures.

\paragraph{Ambiguous Data Model Description.}
Schema documentation provides unclear, misleading, or insufficient column descriptions that prevent accurate interpretation of intended semantics. For example, a column described as ``total amount'' might ambiguously refer to pre-tax totals, post-tax totals, or totals including shipping---with ground truth assuming one interpretation without specification.

\paragraph{Evaluation False Positives.}
Agent outputs and ground truth are semantically equivalent but flagged as mismatched due to evaluation methodology limitations. We identified six subcategories:
\begin{itemize}
    \item \textbf{Actual Match:} Results are functionally identical; discrepancies stem from evaluation script behavior (e.g., floating-point comparison tolerance).
    \item \textbf{Format Mismatch:} Logically correct SQL produces results in alternative valid formats (e.g., 0.15 vs.\ 15\% for percentages).
    \item \textbf{Duplicated Rows in Ground Truth:} Agent SQL is correct; mismatches arise from erroneous duplicate rows in ground truth data.
    \item \textbf{NULL Representation:} Agent uses different NULL handling (e.g., \texttt{COALESCE(column, 0)}) when schema documentation does not specify behavior.
    \item \textbf{Row Ordering:} SQL logic is correct but result ordering differs due to unspecified \texttt{ORDER BY} requirements.
    \item \textbf{Other:} Miscellaneous formatting or representation differences.
\end{itemize}

\paragraph{Ground Truth Calculation Errors.}
Ground truth contains values that cannot be derived from any reasonable query over the available schema:
\begin{itemize}
    \item Values do not match any standard interpretation of the task specification.
    \item Arbitrary or undocumented business rules appear to have been applied.
    \item Agent results achieve near-perfect matches (e.g., 99.8\%) but fail due to anomalous rows with no reasonable explanation.
\end{itemize}

\subsection{Error Distribution}

Table~\ref{tab:error_distribution} presents the distribution of errors across categories for all 660 analyzed columns. Nearly one-third (33.0\%) of all column mismatches are benchmark-attributable rather than genuine agent failures.

At the task level, we observe even more striking patterns. Among the 81 tasks that failed transformation evaluation:

\begin{itemize}
    \item 17.3\% (14 tasks) contain \textbf{only agent-attributable errors}
    \item 14.8\% (12 tasks) contain \textbf{only benchmark-attributable errors}
    \item 67.9\% (55 tasks) contain \textbf{both error types}
\end{itemize}

A total of 67 tasks (82.7\% of failed tasks) contain at least one benchmark-attributable error. These flaws misdirect the community by penalizing correct behavior. We highlight the three benchmark-attributable categories with representative examples (full details in the Appendix):

\paragraph{Evaluation False Positives (23.6\%)}
These occur when an agent's output is semantically correct but is flagged as a failure due to rigid evaluation scripts. A representative case is the \texttt{GNP\_GROWTH\_RATE} column in the \texttt{world} database, described simply as \emph{``The GNP growth rate of the country.''}\ The word ``rate'' naturally suggests a decimal fraction, and the agent computes exactly that (e.g., \texttt{0.0441} for Aruba), yet the ground truth expects the equivalent percentage (\texttt{4.41}). Every one of the 178 non-null values is exactly 100$\times$ smaller than the ground truth---the underlying formula is correct, only the output scale differs. This single formatting choice causes a 0\% match rate (Appendix~\ref{sec:fp_example}). Such false positives force developers to over-engineer their models to guess arbitrary formatting conventions rather than optimizing for correct logic.

\paragraph{Ambiguous Data Model Descriptions (4.8\%)}
These arise when column specifications are underspecified, admitting multiple valid interpretations. For instance, the \texttt{NUM\_WON\_2010} column in the \texttt{european\_football\_2} database asks for ``the number of matches the team won in the 2010 season.'' In football, seasons span two calendar years, so ``2010 season'' could refer to the \texttt{2009/2010} season, the \texttt{2010/2011} season, or calendar year 2010. The agent selects \texttt{2009/2010}---a defensible interpretation---yet the ground truth corresponds to \texttt{2010/2011}, yielding only $\sim$40\% agreement (Appendix~\ref{sec:ambiguous_example}). We deliberately preserve such ambiguities in ELT-Bench-Verified, as resolving underspecified requirements is a capability agents should be evaluated on, not shielded from. 

\paragraph{Ground Truth Calculation Errors (4.5\%)}
These are non-mitigable errors where the ground-truth data contains values that cannot be derived from any reasonable query over the available schema. For example, in the \texttt{codebase\_community} database, a column defined as ``set to 1 if the post was posted by the most influential user'' has ground truth values that match \emph{only} when ``most influential'' is operationalized as users with the specific reputation score of 995---ranking 306th out of all users---while users with reputations exceeding 87{,}000 are labeled as non-influential. This undocumented and counterintuitive rule cannot be deduced from the specification. We tested five standard operationalizations---highest reputation, most posts, most badges, most profile views, and most upvotes---all far more defensible, yet none reproduce the ground truth (Appendix~\ref{sec:gt_example}). Evaluating agents against such targets fundamentally breaks the utility of the benchmark.

Figure~\ref{fig:simple} illustrates the overall pipeline outcomes, while Figure~\ref{fig:cover} shows the error type distribution among failed transformation tasks.


\section{ELT-Bench-Verified}
\label{sec:verified}

Based on our audit findings, we construct ELT-Bench-Verified---a corrected version of the original benchmark. This section presents the corrections applied by the Corrector pipeline (Section~\ref{sec:corrector}) and the validation experiments.

\subsection{Corrections Applied}

\paragraph{Evaluation Script Refinements.}
We modified the transformation stage evaluation script to address false positive categories identified in our taxonomy:
\begin{itemize}
    \item Standardized boolean comparisons (``true''/``false'' vs.\ 0/1)
    \item Implemented appropriate floating-point comparison tolerances
    \item Added format normalization for percentages and decimals
    \item Handled NULL representation equivalences
    \item Made row-order comparisons order-insensitive when no ordering is specified
\end{itemize}

\paragraph{Ground Truth Column Removal.}
For the 30 columns (spanning 24 tasks) classified as Ground Truth Calculation Errors, an independent validation study (Section~\ref{sec:corrector_validation}) revealed that human experts themselves disagree on the correct interpretation, with an average pairwise agreement of only 57.8\%. Since no reliable correction can be established, we remove these 30 columns from the benchmark entirely---amounting to just 1.2\% of all columns (30 out of 2{,}494)---while retaining all remaining columns in each affected data model.

\paragraph{Preservation of Ambiguity.}
We deliberately did not modify task descriptions for the 32 columns classified as Ambiguous Data Model Description errors. Handling specification ambiguity is a capability that AI agents should possess---potentially through specialized disambiguation agents---and this capability should be evaluated rather than circumvented through benchmark simplification.

\begin{table}[tb]
\centering
\caption{Agent performance before and after benchmark correction. Both evaluations use SWE-Agent with Claude Sonnet 4.5.}
\label{tab:verified_results}
\begin{tabular}{lccc}
\toprule
\textbf{Benchmark} & \textbf{SRDEL} & \textbf{SRDT} & \textbf{Models Passed} \\
\midrule
ELT-Bench (original) & 96\% & 22.66\% & 46/203 \\
ELT-Bench-Verified & 96\% & 32.51\% & 66/203 \\
\bottomrule
\end{tabular}
\end{table}

\subsection{Validation}

To validate that our corrections have a measurable effect on evaluation outcomes and are not overfit to a single agent configuration, we conduct three experiments.

\subsubsection{End-to-End Re-evaluation}
\label{sec:e2e}

Table~\ref{tab:verified_results} presents the results of re-evaluating SWE-Agent with Claude Sonnet 4.5 on ELT-Bench-Verified using the same experimental setup described in Section~\ref{sec:reeval}. SRDEL remains at 96\%, as the corrections target only the transformation stage. SRDT increases from 22.66\% to 32.51\%---a 9.85 percentage point gain (43.5\% relative improvement), with 20 additional data models now passing evaluation. Importantly, the denominator remains 203 data models: the 30 removed columns are distributed across data models that retain other columns, so no data model is eliminated from the benchmark. The improvement reflects two effects: (i)~correcting evaluation false positives that previously penalized correct agent outputs, and (ii)~removing unreliable ground truth columns that no agent could match correctly.

\subsubsection{Ablation of Correction Strategies}
\label{sec:ablation}

To disentangle the contributions of each correction strategy, we evaluate SWE-Agent with Claude Sonnet 4.5 under three configurations, using the same experimental setup described in Section~\ref{sec:reeval}: (i)~\emph{evaluation script refinement only}, which applies the patched evaluation logic while retaining the original ground truth and all columns; (ii)~\emph{GT column removal only}, which removes the 30 columns with unreliable ground truth while retaining the original evaluation script; and (iii)~\emph{ELT-Bench-Verified}, which combines both corrections. Table~\ref{tab:ablation} presents the results.

\begin{table}[tb]
\centering
\caption{Ablation of correction strategies. All evaluations use SWE-Agent with Claude Sonnet 4.5. The baseline is the original ELT-Bench evaluation.}
\label{tab:ablation}
\begin{tabular}{lcc}
\toprule
\textbf{Configuration} & \textbf{SRDT} & \textbf{Models Passed} \\
\midrule
ELT-Bench (original) & 22.66\% & 46/203 \\
+ Evaluation script refinement only & 30.05\% & 61/203 \\
+ GT column removal only & 24.63\% & 50/203 \\
ELT-Bench-Verified (both) & 32.51\% & 66/203 \\
\bottomrule
\end{tabular}
\end{table}

Evaluation script refinements account for the majority of the improvement (15 additional models), while GT column removal contributes a smaller but complementary gain (4 additional models). The combined correction yields 20 additional models---one more than the sum of individual gains (15+4=19)---indicating that at least one data model contains both evaluation false positives and ground truth errors, requiring both corrections simultaneously to pass.

\subsubsection{Multi-Agent Validation}
\label{sec:multiagent}

To demonstrate that the corrections are not overfit to a single agent, we evaluate a second agent\\configuration---a Baseline agent implemented in LangGraph~\cite{langgraph} using the ReAct~\cite{react} agentic paradigm, paired with Claude Sonnet 4.5---on both the original ELT-Bench and ELT-Bench-Verified. Table~\ref{tab:multiagent} presents the results. Both agents show substantial improvement on ELT-Bench-Verified, confirming that the identified errors were genuine benchmark issues rather than artifacts of one agent's behavior. Notably, both agents converge to the same SRDT (32.51\%, 66/203 models) on the corrected benchmark despite differing on the original (22.66\%, 46/203 models for SWE-Agent vs.\ 20.20\%, 41/203 models for ReAct). However, the two agents arrive at the same 32.51\% SRDT via different paths: ReAct achieves higher SRDEL (98\% vs.\ 96\%) and matches more columns overall (1{,}986 vs.\ 1{,}854), yet coincidentally passes the same number of data models (66/203). As shown in Appendix~\ref{sec:multiagent_detail}, the 66 passing models are not identical across the two agents. The fact that both agents improve substantially on ELT-Bench-Verified---despite using different agentic strategies---provides evidence that the corrections address genuine benchmark defects rather than artifacts of a single agent's behavior.

\begin{table}[tb]
\centering
\caption{Multi-agent validation. Performance of two agent configurations on the original and corrected benchmark.}
\label{tab:multiagent}
\begin{tabular}{llccc}
\toprule
\textbf{Agent} & \textbf{Benchmark} & \textbf{SRDEL} & \textbf{SRDT} \\
\midrule
\makecell[l]{SWE-Agent\\w/ Claude Sonnet 4.5} & ELT-Bench & 96\% & 22.66\% \\
& ELT-Bench-Verified & 96\% & 32.51\% \\
\midrule
\makecell[l]{Baseline (ReAct)\\w/ Claude Sonnet 4.5} & ELT-Bench & 98\% & 20.20\% \\
& ELT-Bench-Verified & 98\% & 32.51\% \\
\bottomrule
\end{tabular}
\end{table}

\section{Related Work}

\paragraph{ELT Pipelines and Automation.}
The evolution from ETL to ELT reflects a broader shift toward cloud-native data architectures~\cite{ETLhistory}, where elastic compute in platforms such as Snowflake~\cite{snowflake} enables transformations to execute directly within the warehouse. This paradigm preserves raw data fidelity, supports schema-on-read flexibility, and reduces the operational overhead of maintaining dedicated transformation infrastructure. However, constructing ELT pipelines remains labor-intensive, requiring expertise in source-specific extraction configuration (e.g., Airbyte~\cite{airbyte}), data loading orchestration, and SQL-based transformation logic (e.g., dbt~\cite{dbt}). Recent work has explored AI-driven automation of individual pipeline stages, but standardized evaluation of these capabilities has lagged behind system development.

\paragraph{Benchmarks for Data Engineering.}
Existing benchmarks for data-related AI tasks predominantly focus on text-to-SQL evaluation. Widely adopted benchmarks such as BIRD~\cite{bird} and Spider 2.0~\cite{spider2} have been central to measuring progress in SQL generation, but they evaluate only the transformation stage in isolation, assuming data is already available in queryable form. The extraction, loading, and tool configuration stages that precede transformation in real ELT workflows fall outside their scope. TPC-DI~\cite{TPCDI}, the industry-standard benchmark for data integration, covers the full ETL workflow but was designed for traditional data integration systems rather than AI agents and follows the ETL (transform-before-load) paradigm rather than modern ELT.

ELT-Bench~\cite{eltbench} addresses this gap as the first benchmark specifically designed to evaluate AI agents on end-to-end ELT pipeline construction. By requiring agents to configure extraction from heterogeneous sources, orchestrate loading into Snowflake, and write dbt transformation models, ELT-Bench evaluates capabilities that no prior benchmark captures. Our work builds on ELT-Bench by auditing its data quality and releasing a corrected version.

\paragraph{Annotation Errors and Benchmark Auditing.}
A growing body of literature has identified systematic quality issues in widely used benchmarks across NLP, software engineering, and data management, including annotation inconsistencies, ambiguous task specifications, and flawed ground truth labels~\cite{pourreza2023evaluating, swebenchplus}. In the text-to-SQL domain, prior analyses have reported annotation errors in BIRD~\cite{birdnoise}, and a recent comprehensive audit demonstrated that error rates in BIRD Mini-Dev and Spider 2.0-Snow can exceed 50\%, substantially distorting leaderboard rankings~\cite{jin2025pervasive}.

To address similar challenges, prior work has combined automated validation techniques with expert review, using generated test cases, oracle construction, and execution-based validation to strengthen evaluation rigor. Our methodology follows this paradigm: we employ automated LLM-based root cause analysis at scale (Phase~2) but require comprehensive human validation of all findings (Phase~3). Our results extend the observation of pervasive benchmark-attributable errors from text-to-SQL settings to the more complex domain of end-to-end ELT pipeline evaluation.

\section{Discussion}

\subsection{The Evolving Landscape of ELT Automation}

The progression from 1\% to 22.66\% to 32.51\% SRDT reflects two compounding factors that obscured agent capabilities: the use of an older model in the original evaluation, and benchmark quality issues. The first gap is attributable to model improvement (with the agent framework held constant), while the second is attributable entirely to benchmark correction. At 96\% SRDEL, the extraction and loading stage is largely solved, suggesting that future benchmark development should focus on more challenging pipeline stages.

\subsection{Implications for Benchmark Design}

Our findings extend recent observations of pervasive annotation errors in text-to-SQL benchmarks~\cite{jin2025pervasive} to end-to-end ELT pipeline evaluation. These results reinforce the need for systematic quality auditing as a standard practice in benchmark development, particularly for agentic tasks where evaluation involves multi-step pipelines and execution-based correctness checking.

\subsection{Implications for Agentic ELT Systems}
\label{sec:implications_agentic}

Our error taxonomy provides a detailed map of where current AI agents succeed and fail in ELT pipeline construction, offering actionable guidance for practitioners building agentic data engineering systems.

\paragraph{What agents can already do.}
The extraction and loading stage is effectively solved: at 96\% SRDEL, current models can reliably interpret connection configurations, generate Terraform definitions for Airbyte connectors, and orchestrate data loading into Snowflake with minimal failure. For straightforward transformation tasks---those involving direct column mappings, simple aggregations, and well-specified schemas---agents also perform reliably, as evidenced by the 1,854 (SWE-Agent) and 1,986 (ReAct) columns matched correctly out of 2,202 eligible. Practitioners can confidently deploy agents for the EL stage and for transformation tasks with clear, unambiguous specifications.

\paragraph{Where agents struggle: recurring error patterns.}
Our taxonomy reveals that agent failures are heavily concentrated in two categories: \emph{Flawed SQL Logic} (33.5\% of all errors) and \emph{JOIN Type Errors} (23.8\%), which together account for 57.3\% of all agent-attributable failures. Importantly, these are not failures of task comprehension---agents generally understand \emph{what} to compute---but rather failures of SQL construction. By contrast, domain knowledge gaps account for only 2.4\% of errors, indicating that semantic understanding is rarely the bottleneck. A detailed examination of the 660 column-level errors reveals several recurring patterns:

\begin{itemize}
    \item \textbf{Aggregation with ties.} A frequent source of errors is the use of \texttt{LIMIT 1} to retrieve a superlative (e.g., ``the film with the highest replacement cost''). When multiple entities share the same maximum value, \texttt{LIMIT 1} returns an arbitrary result. This pattern recurs across databases and accounts for a notable share of Flawed SQL Logic errors.

    \item \textbf{NULL handling assumptions.}
    Agents often make incorrect assumptions about NULL semantics: replacing NULLs with zeros when the ground truth preserves them, excluding NULLs from ratio computations when they should be included, or using \texttt{COALESCE(..., 'Null')} (a string literal) instead of preserving actual NULL values. In some cases, the same database uses different NULL conventions across columns, making consistent handling particularly challenging.

    \item \textbf{String fidelity and data-value assumptions.}
    Agents frequently apply implicit normalization---trimming whitespace, altering capitalization, or reformatting values---when the ground truth expects source data to be preserved exactly. We observe whitespace-related errors across 12 columns spanning 8 databases. Similarly, agents assume standard representations for real-world entities that differ from the actual data (e.g., filtering for \texttt{`USA'} when the data uses \texttt{`United States'}, or missing a brand name due to punctuation differences).
\end{itemize}

\paragraph{Strategies for improving agentic ELT systems.}
The concentration of errors in SQL mechanics rather than task comprehension suggests several high-leverage strategies:

\begin{itemize}
    \item \textbf{Execution-based self-verification.} Agents should execute their generated queries and verify that output schema, row counts, and value distributions match expectations. Our Auditor pipeline demonstrates feasibility: the LLM agent self-validated corrected SQL for 630 of 660 columns. Agents should specifically check for unexpected row count changes after JOINs (a signal for incorrect join types, which alone account for 23.8\% of failures) and verify that superlative queries handle ties correctly.

    \item \textbf{Data profiling before query generation.} Many errors stem from agents assuming data formats without inspecting actual values. A preliminary profiling step---sampling column values to detect whitespace patterns, NULL rates, string formats, and numeric representations---could prevent a substantial class of errors (e.g., revealing that a boolean column uses \texttt{0/1} rather than \texttt{True/False}, or that country names use full forms rather than abbreviations).

    \item \textbf{Multi-agent portfolios.} SWE-Agent and ReAct share only 14 of their fully correct databases, with each exclusively solving additional tasks the other cannot (Section~\ref{sec:multiagent}). This complementarity suggests that ensemble approaches---running multiple agents and selecting the best output per task---could substantially improve coverage.

    \item \textbf{Human-in-the-loop for ambiguity resolution.} Ambiguous specifications account for 4.8\% of column errors, where agents select a defensible but incorrect interpretation (e.g., ``the most common film genre'' without tie-breaking behavior). Multi-agent debate architectures, where several agents independently propose interpretations and a human resolves disagreements, are a promising direction for production ELT systems.
\end{itemize}

\paragraph{What remains hard.}
Even on ELT-Bench-Verified, 67.5\% of data models still fail. The hardest cases involve the interplay of multiple error types within a single task---a column requiring both a correct multi-table join \emph{and} proper NULL handling \emph{and} data-aware string matching. These compounding difficulties, combined with domain-specific business logic and long chains of dependent SQL operations, suggest that future progress will require advances in structured reasoning over relational schemas and data-aware query generation.

\subsection{Limitations}
\label{sec:limitations}
Our analysis has several limitations. First, our Auditor-Corrector pipeline was applied to the failed tasks of a single agent configuration (SWE-Agent with Claude Sonnet 4.5). Different agents may fail on different tasks, potentially revealing additional benchmark-attributable errors not captured by our audit. While we validate the corrected benchmark with a second agent (Section~\ref{sec:multiagent}), applying the full Auditor pipeline to the failures of multiple agents would provide broader coverage; however, as discussed in Section~\ref{sec:reeval}, each full evaluation run requires multiple days and hundreds of dollars in API costs, making exhaustive multi-agent auditing prohibitively expensive. Second, the primary categorization of all 660 columns into our error taxonomy was performed by a single data engineer. An inter-annotator agreement study on a random sample of 50 columns yields Fleiss' $\kappa = 0.851$ (almost perfect) for the high-level agent-vs.-benchmark attribution and $\kappa = 0.755$ (substantial) for the exact 14-category classification, indicating high reproducibility. However, the study covers a sample rather than the full set of 660 columns, and edge cases in the complete dataset may exhibit lower consistency. Third, our error attribution required judgment calls in ambiguous cases. In particular, the boundary between ``Ambiguous Data Model Description'' (preserved in the benchmark) and ``Evaluation False Positive'' (corrected) involves a judgment about whether the specification is genuinely unclear or merely admits representational variation---reasonable annotators might draw this line differently.

\section{Conclusion}

We presented a comprehensive reassessment of ELT-Bench revealing that model improvement (SRDEL: 37\%$\to$96\%, SRDT: 1\%$\to$22.66\%) and benchmark quality issues (SRDT: 22.66\%$\to$32.51\% after correction) jointly led the original evaluation to substantially underestimate agent capabilities. Our systematic audit found that 82.7\% of failed tasks contain benchmark-attributable errors, and we release ELT-Bench-Verified as a corrected benchmark. While transformation remains challenging---67.5\% of data models still fail---our error taxonomy and analysis of recurring failure patterns provide actionable guidance for building more effective agentic ELT systems.


\appendix

\section{Illustrative Examples}

\subsection{Ground Truth Error}
\label{sec:gt_example}

We illustrate the nature of ground truth calculation errors with a concrete example from the \texttt{codebase\_community} database. The column \texttt{POSTED\_BY\_MOST\_INFLUENTIAL\_USER} is described as: \emph{``Set to 1 if the post was posted by the most influential user, otherwise, set to 0.''}

The term ``most influential user'' is inherently ambiguous---it could reasonably refer to the user with the highest reputation, the most posts, the most upvotes, or other engagement metrics. Table~\ref{tab:gt_example_interpretations} lists five standard interpretations, each producing a different user and a different match rate against the ground truth.

\begin{table}[tb]
\centering
\caption{Alternative interpretations of ``most influential user'' tested against the ground truth (91{,}966 posts total). No standard interpretation reproduces the GT; only an undocumented filter (\texttt{Reputation\,=\,995}) achieves 100\%.}
\label{tab:gt_example_interpretations}
\begin{tabular}{llr}
\toprule
\textbf{Interpretation} & \textbf{Selected User(s)} & \textbf{Match} \\
\midrule
Most posts          & User 805 (1{,}720 posts)        & 98.07\% \\
Highest reputation  & User 919 (rep.\ 87{,}393)       & 98.63\% \\
Most badges         & User 919                         & 98.63\% \\
Most profile views  & User 919 (20{,}932 views)       & 98.63\% \\
Most upvotes        & User 1036                        & 99.74\% \\
\midrule
\texttt{Reputation = 995} & Users 3432 \& 8165        & \textbf{100\%} \\
\bottomrule
\end{tabular}
\end{table}

The ground truth marks exactly 56 posts as being from the ``most influential'' user. These 56 posts belong to two users (3432 and 8165), both with a reputation score of exactly 995---placing them at rank 306--307 out of all users (99.24th percentile). Meanwhile, users with reputations of 65{,}272 and 87{,}393 are labeled as non-influential. Table~\ref{tab:gt_example_users} highlights this contrast.

\begin{table}[tb]
\centering
\caption{User comparison for the \texttt{POSTED\_BY\_MOST\_INFLUENTIAL\_USER} column. The GT flags only users with \texttt{Reputation\,=\,995} as ``most influential,'' ignoring users with far higher reputation, post counts, and engagement metrics.}
\label{tab:gt_example_users}
\begin{tabular}{rrrrrr}
\toprule
\textbf{User} & \textbf{Rep.} & \textbf{Posts} & \textbf{UpVotes} & \textbf{Views} & \textbf{GT=1} \\
\midrule
805   & 65{,}272 & 1{,}720 & 7{,}035  & 5{,}680  & 0  \\
919   & 87{,}393 & 1{,}204 & 11{,}273 & 20{,}932 & 0  \\
3432  & 995      & 22      & 79       & 82       & 22 \\
8165  & 995      & 34      & 66       & 81       & 34 \\
\bottomrule
\end{tabular}
\end{table}

While the column description is genuinely ambiguous---a reasonable point of discussion---the ground truth error is unambiguous: no standard definition of ``influence'' would select users ranked 306th by reputation over users ranked 1st or 2nd. The GT was evidently generated using an undocumented filter (\texttt{WHERE Reputation = 995}) that bears no relationship to the stated specification. This column is one of the 30 removed from ELT-Bench-Verified.

\subsection{Evaluation False Positive}
\label{sec:fp_example}

We illustrate evaluation false positives with the \texttt{GNP\_GROWTH\_RATE} column in the \texttt{world} database, described as: \emph{``The GNP growth rate of the country.''}

The agent applies the standard growth-rate formula $(\mathit{GNP} - \mathit{GNP}_{\mathit{old}}) / \mathit{GNP}_{\mathit{old}}$, returning a decimal fraction (e.g., 0.0441 for a 4.41\% growth). The ground truth stores the same quantity as a percentage (4.41). The two representations are mathematically equivalent---every value differs by exactly a factor of 100---yet the evaluation script performs an exact string comparison, yielding a 0\% match rate across all 178 non-null countries.

\begin{table}[tb]
\centering
\caption{Format mismatch for \texttt{GNP\_GROWTH\_RATE}. The agent's formula is correct; only the output scale differs. All 178 non-null values exhibit an exact 100$\times$ ratio.}
\label{tab:fp_example}
\begin{tabular}{lrrr}
\toprule
\textbf{Country} & \textbf{GT (\%)} & \textbf{Pred (decimal)} & \textbf{Ratio} \\
\midrule
Aruba     &   4.41  & 0.0441  & 100.0 \\
Angola    & $-$16.73 & $-$0.1673 & 100.0 \\
Albania   &  28.20  & 0.2820  & 100.0 \\
UAE       &   3.04  & 0.0304  & 100.0 \\
Argentina &   5.24  & 0.0524  & 100.0 \\
\bottomrule
\end{tabular}
\end{table}

Table~\ref{tab:fp_example} shows representative rows. The ratio between ground truth and predicted values is exactly 100.0 for every row (standard deviation $\approx 0$). The column description uses the word ``rate,'' which in standard mathematical usage denotes a ratio (0--1), not a percentage (0--100). The agent's interpretation is arguably more faithful to the specification than the ground truth, yet it receives a 0\% score. Multiplying by 100 yields a perfect 178/178 match, confirming that the mismatch is purely representational.

\subsection{Ambiguous Data Model Description}
\label{sec:ambiguous_example}

We illustrate ambiguous data model descriptions with the\\\texttt{NUM\_WON\_2010} column in the \texttt{european\_football\_2} database, described as: \emph{``The number of matches the team won in the 2010 season, with NULL values replaced with 0.''}

The term ``2010 season'' is inherently ambiguous in a football context, where seasons span two calendar years. The agent interprets it as the \texttt{2009/2010} season (August 2009--May 2010), a defensible reading. However, the ground truth values correspond to a different interpretation---either the \texttt{2010/2011} season or a calendar-year 2010 filter. Table~\ref{tab:ambiguous_example} shows the resulting mismatches.

\begin{table}[tb]
\centering
\caption{Season interpretation ambiguity for \texttt{NUM\_WON\_2010}. The agent uses the \texttt{2009/2010} season; the ground truth implies a different interpretation. Only $\sim$40\% of teams match under the agent's reading.}
\label{tab:ambiguous_example}
\begin{tabular}{lrrr}
\toprule
\textbf{Team} & \textbf{GT} & \textbf{Pred} & \textbf{Diff} \\
\midrule
Ruch Chorzow          & 10 & 16 & $-$6 \\
Jagiellonia Bialystok & 14 & 11 & $+$3 \\
Lech Poznan           & 15 & 19 & $-$4 \\
P.\ Warszawa          & 10 &  9 & $+$1 \\
Cracovia              &  6 &  9 & $-$3 \\
S.C.\ Olhanense       &  7 &  5 & $+$2 \\
FC Pacos de Ferreira  &  8 &  8 &  0 \\
SM Caen               &  4 &  0 & $+$4 \\
\bottomrule
\end{tabular}
\end{table}

The specification does not clarify whether ``2010 season'' refers to the season starting in 2010 (\texttt{2010/2011}), the season ending in 2010 (\texttt{2009/2010}), or matches played in the calendar year 2010. The agent's choice of \texttt{2009/2010} is a reasonable default, yet it produces only $\sim$40\% agreement with the ground truth. Testing the \texttt{2010/2011} season yields a 100\% match, confirming that the mismatch stems entirely from an underspecified temporal reference rather than flawed SQL logic.

\section{Multi-Agent Validation Detail}
\label{sec:multiagent_detail}

Table~\ref{tab:multiagent_detail} provides a detailed comparison of SWE-Agent and Baseline (ReAct) on ELT-Bench-Verified. Although both agents pass 66 data models (32.51\% SRDT), the sets of fully correct databases differ. SWE-Agent achieves perfect scores on 18 databases, while ReAct achieves perfect scores on 15. The two agents share 14 fully correct databases; SWE-Agent exclusively solves 4 additional databases (\texttt{instagram\_business}, \texttt{law\_episode}, \texttt{linkedin}, \texttt{menu}), while ReAct exclusively solves 1 (\texttt{bike\_share\_1}). This confirms that the identical SRDT is coincidental: SWE-Agent's correct models are more concentrated within fewer databases, while ReAct's are more evenly distributed across a broader set of tasks.

\begin{table}[tb]
\centering
\caption{Detailed comparison of agent performance on ELT-Bench-Verified. Both agents pass 66/203 data models but differ in column-level behavior and fully correct databases.}
\label{tab:multiagent_detail}
\begin{tabular}{lcc}
\toprule
\textbf{Metric} & \textbf{SWE-Agent} & \textbf{ReAct} \\
\midrule
SRDEL & 96\% & 98\% \\
SRDT & 32.51\% & 32.51\% \\
Eligible data models & 194 & 199 \\
Correct data models & 66 & 66 \\
Matched columns & 1{,}854 & 1{,}986 \\
Unmatched columns & 348 & 411 \\
Fully correct databases & 18 & 15 \\
\midrule
Shared fully correct DBs & \multicolumn{2}{c}{14} \\
SWE-Agent only & \multicolumn{2}{c}{4} \\
ReAct only & \multicolumn{2}{c}{1} \\
\bottomrule
\end{tabular}
\end{table}

\section*{Use of AI}
The authors used AI-assisted writing tools (e.g., large language models) for grammar checking and improving the clarity of the manuscript. All content was written and reviewed by the authors, who take full responsibility for the work.

\bibliographystyle{ACM-Reference-Format}
\bibliography{sample}

\end{document}